\documentclass[runningheads]{llncs}
\usepackage{graphicx}
\usepackage{amsmath}
\usepackage{amssymb}
\usepackage{float}
\usepackage{graphicx}
\usepackage{amsmath,amssymb,amsfonts}
\usepackage{threeparttable}
\usepackage{xcolor}
\usepackage{multirow}
\usepackage{color}
\usepackage{float}
\usepackage{booktabs}
\usepackage{bigstrut}
\usepackage{balance}
\usepackage{algorithm}
\usepackage{xcolor}
\usepackage{color, colortbl}
\usepackage[noend]{algpseudocode}
\usepackage{cite}
\usepackage{enumitem}
\setlist{nosep}
\usepackage{diagbox}
\definecolor{mygray}{gray}{.85}
\definecolor{reda}{RGB}{255,0,0}
\definecolor{redb}{RGB}{217,148,143}
\definecolor{myyellow}{RGB}{190,144,0}
\definecolor{mygreen}{RGB}{0,136,51}
\definecolor{myblue}{RGB}{0,102,204}
\begin{document}
\title{Few-Shot Segmentation via \\ Rich Prototype Generation and \\ Recurrent Prediction Enhancement}
%
%
\author{Hongsheng Wang, Xiaoqi Zhao, Youwei Pang and Jinqing Qi{*}}

\authorrunning{H. Wang \emph{et al}.}
%
\institute{Dalian University of Technology, Dalian, China\\
\email{\{wanghongsheng,zxq,lartpang\}@mail.dlut.edu.cn,jinqing@dlut.edu.cn}}
\maketitle              
\begin{abstract}
Prototype learning and decoder construction are the keys for few-shot segmentation. However, existing methods use only a single prototype generation mode, which can not cope with the intractable problem of objects with various scales. Moreover, the one-way forward propagation adopted by previous methods may cause information dilution from registered features during the decoding process.
In this research, we propose a rich prototype generation module (RPGM) and a recurrent prediction enhancement module (RPEM) to reinforce the prototype learning paradigm and build a unified memory-augmented decoder for few-shot segmentation, respectively.
Specifically, the RPGM combines superpixel and K-means clustering to generate rich prototype features with complementary scale relationships and adapt the scale gap between support and query images. 
The RPEM utilizes the recurrent mechanism to design a round-way propagation decoder. In this way, registered features can provide object-aware information continuously.
Experiments show that our method consistently outperforms other competitors on two popular benchmarks PASCAL-${{5}^{i}}$ and COCO-${{20}^{i}}$.

{\let\thefootnote\relax\footnote{{{*}Corresponding author}}}
\end{abstract}

\section{Introduction}
\begin{figure}[!t]
    \centering
    \includegraphics[width=\linewidth]{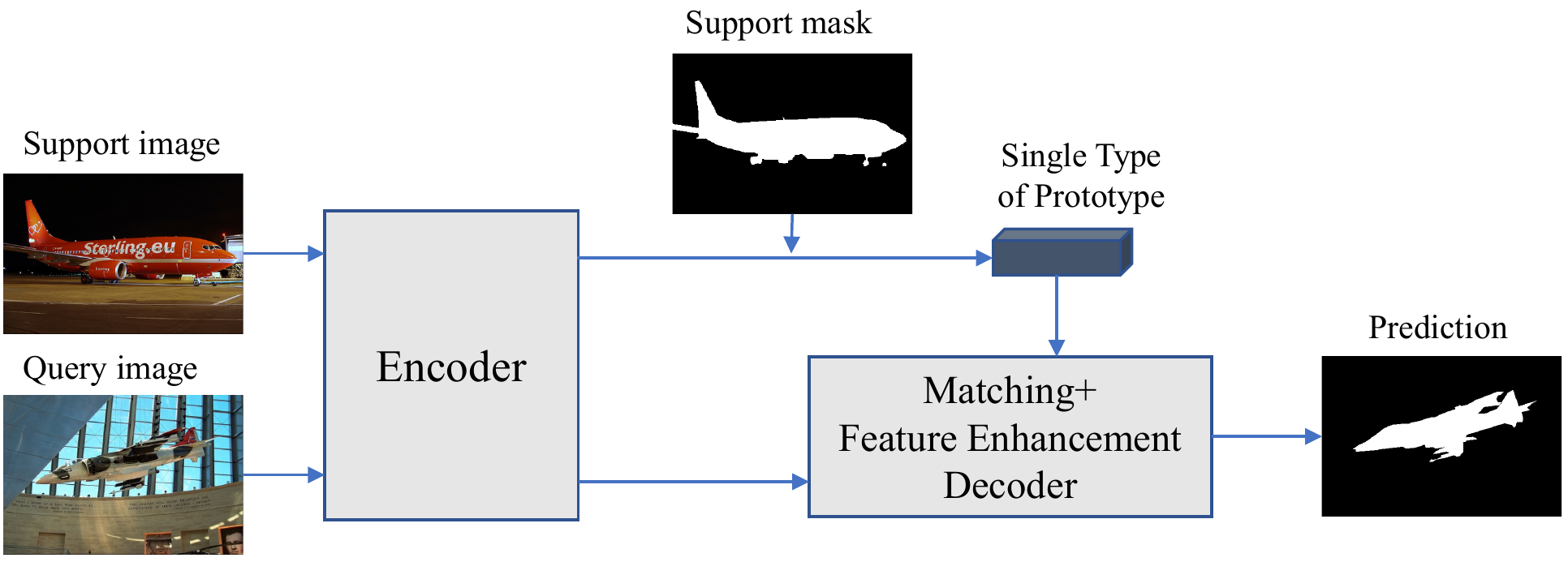}
    \caption{
	A popular few-shot segmentation architecture.
	}
	\label{introdiction}
\end{figure}

In recent years, with the use of deep neural networks and large-scale datasets, 
significant progress has been made in fully-supervised semantic segmentation~\cite{deeplab,E-D,ccnet,improving,pyramid}.
However, the labor cost of acquiring a large number of labeled datasets is very expensive.
To address this challenge, the few-shot segmentation task~\cite{OLS} has been proposed. 
It aims to segment a new object class with only one or a few annotated examples, which is agnostic to the network at the training phase.
Most methods adopt the general structure as shown in Fig.~\ref{introdiction}.
Prototype learning and decoder construction play an important role in few-shot segmentation.
The prototype represents only object-related features and does not contain any background information. Some efforts~\cite{panet,canet,FWB,PL,PMM} investigate different prototype feature generation mechanisms to provide an effective reference for query images. 
Both CANet~\cite{canet} and PFENet~\cite{PFEnet} generate a single prototype by the masked average pooling operation to represent all features in the foreground of the support image.
SCL~\cite{SCL} uses a self-guided mechanism to produce an auxiliary feature prototype.
ASGNet~\cite{ASGNet} is proposed to split support features adaptively into several feature prototypes and select the most relevant prototype to match the query image.
However, the aforementioned methods all adopt a single approach to construct prototype features and ignore complex scale differences between support images and query images, which may introduce scale-level interference for the subsequent similarity measure.
The decoder can finish the feature aggregation and transfer them into the task-required mode.
Nevertheless, many methods~\cite{PFEnet,SCL,ASGNet,pgnet,PPNet} focus on designing the feature enrichment module or applying the multi-scale structure (e.g. ASPP~\cite{aspp}) directly to aggregate the query features through a one-way forward propagation and obtain the final prediction results.
This limitation not only makes the semantic information of the probability map generated by mid-level features insufficient, but also results in truly useful features not being adequately utilized due to information dilution.

In response to these challenges, we propose a rich prototype generation module (RPGM) and a recurrent prediction enhancement module (RPEM) to improve the performance for few-shot segmentation.
The RPGM combines two clustering strategies, superpixel and K-means, to generate rich prototype features that are complete representations of the supporting feature information.
Superpixel clustering can generate $N_{s} \in \{1, \dots, N\}$ prototypes depending on the size of the image, 
while K-means clustering generates specific $N_{k}=N$ prototypes regardless of the image size.
The RPEM is a round-way feedback propagation module based on the original forward propagation decoder and is motivated by the recurrent mechanism.
Specifically, it is composed of a multi-scale iterative enhancement (MSIE) module and a query self-contrast enhancement (QSCE) module.
The former produces multi-scale information for the registered features of each stage, while the latter performs the self-contrast operation on query prototype features and then corrects those registered features.
In this way, 
object-aware information can be constantly obtained from the registered features.
In addition, taking into account the parameter-free nature, the proposed RPEM can also be considered as a flexible post-processing technology by using it only during the inference phase. 

Our main contributions can be summarized as follows: 
\begin{itemize}[noitemsep,nolistsep]
	\item For few-shot segmentation, we design two simple yet effective improvement strategies from the perspectives of prototype learning and decoder construction. %
	\item We put forward a rich prototype generation module, which generates complementary prototype features at two scales through two clustering algorithms with different characteristics.
	\item An more efficient semantic decoder is powered by the proposed novel recurrent prediction enhancement module, where multi-scale and discriminative information is adequately propagate to each decoder block.
	\item Extensive experiments on two benchmark datasets demonstrate that the proposed model outperforms other existing competitors under the same metrics. 
\end{itemize}

\section{Related Work}

\noindent \textbf{Semantic Segmentation} is a fundamental computer vision task that aims to accurately predict the label of each pixel. 
Currently, the encoder-decoder architecture~\cite{Segnet,E-D} is widely used.
The encoder extracts high-level semantic features at low resolution, while the decoder progressively recovers the resolution of feature maps to obtain the segmentation mask.
Besides, many semantic segmentation methods adopt the pyramid pooling structure~\cite{psanet,context,ccnet} to capture semantic context from multiple perspectives.
Although these methods achieve good performance, they rely on pixel-level annotation of all classes in the training phase and can not be generalized to those new classes with only a few number of labels.

\noindent \textbf{Few-shot Learning} is proposed to leverage limited prior knowledge to predict new classes. 
Current solutions are mainly based on meta-learning~\cite{memory,mmf,mle} and metric learning~\cite{mnol,lsl,deepemd}. 
Meta-learning aims to obtain a model that can be quickly adapted to new tasks using previous experience, while metric learning models the similarity among objects to generate discriminative representations for new categories. 

\noindent \textbf{Few-shot Segmentation} aims to segment query images containing new categories through utilizing useful information from a small number of labeled data.
PL~\cite{PL} is the first to introduce prototype learning into few-shot segmentation and obtains segmentation results by comparing support prototypes and query features. 
Prototype alignment regularization is used in PANet~\cite{panet}, which encourages mutual guidance between support and query images.
PGNet~\cite{pgnet} introduces a graph attention unit to explore the local similarity between support and query features.
PPNet~\cite{PPNet} moves away from the limitations of the overall prototype and proposes partial perception prototypes that represent fine-grained features.
In PFENet~\cite{PFEnet}, a prior generated by advanced features of the support and query is utilized to guide the segmentation. 
Nevertheless, the above methods can not capture fully geometric information limited by employing a single prototype generation mode.
Once there is a large size gap between the support and query objects, the similarity calculation between them will produce large errors and interfere with the decoder to generate the final prediction.

\begin{figure}[t]
    \centering
    \includegraphics[width=\linewidth]{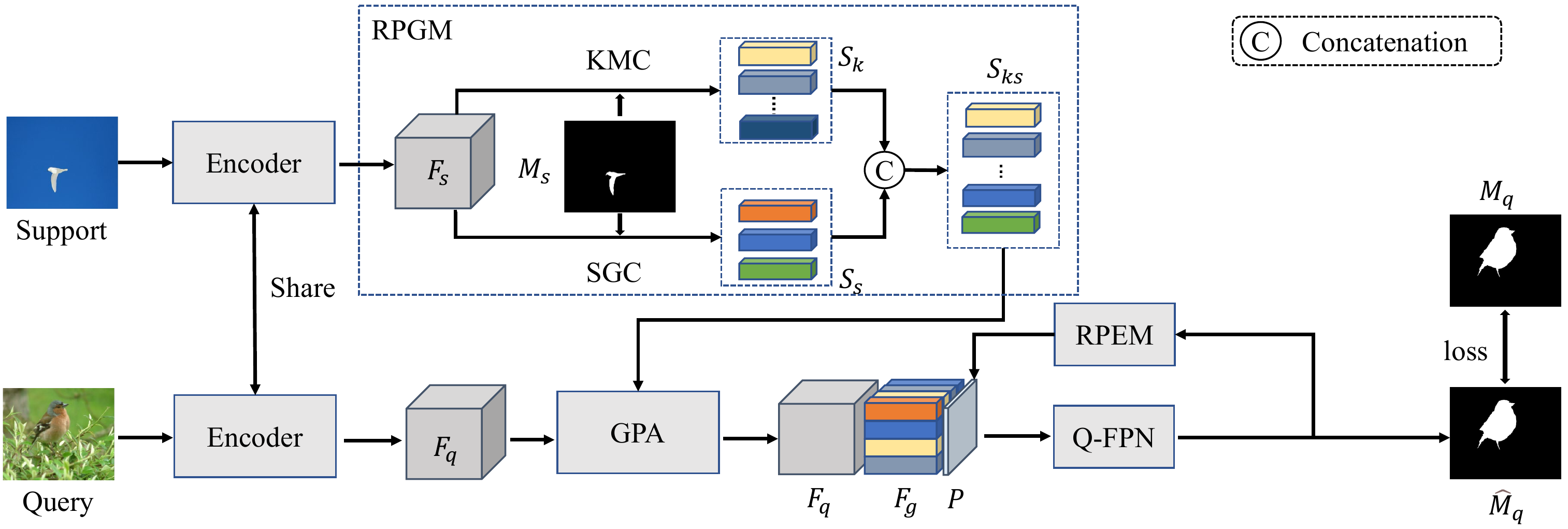}
   \caption{
    	The overall structure of our method with the rich prototype generation module (RPGM) and the recurrent prediction enhancement module (RPEM).
	}
	\label{wang1}
\end{figure}

\section{Problem Definition}

In few-shot segmentation task, the categories in the training and test datasets are disjoint, models are trained on base classes ${{C}_{train}}$ and tested on novel classes ${{C}_{test}}$ in episodes $({{C}_{train}}\cap {{C}_{test}}=\varnothing )$. Each episode consists of a support set $S$ and a query set $Q$ belonging to the same class ${{C}}$.
The support set $S$ has $K$ samples  $S=\{{{S}_{1}}, {{S}_{2}}, \ldots,{{S}_{K}}\}$, and the $i$-th support sample ${{S}_{i}}$ consists of a support image ${{I}_{s}}$ and a pixel-wise annotation ${{M}_{s}}$ with class ${{C}}$.
Also, the query set $Q$ consists simply of a query image ${{I}_{q}}$ and the ground truth ${{M}_{q}}$ indicating the object belonging to class ${{C}}$.
The query-support pair $\{{{I}_{q}},S\}$ forms the input branch of the network and the ground truth ${{M}_{q}}$ of the query image is not available during training, but is used for the evaluation of the query image.

\section{Proposed Approach}
\label{section:NetworkArchitectures}

\subsection{Overall Framework}
\label{sec1}

Our network architecture shown in Fig.~\ref{wang1} uses a strong baseline ASGNet~\cite{ASGNet} to explore the effectiveness of each proposed component. 
First, we feed the support and query images into a shared encoder to extract features. 
And then, the support features are passed through the rich prototype generation module (RPGM) to produce two different representations of the prototype and are matched to the query features by the guided prototype allocation (GPA) module.
Meanwhile, in the GPA module, the cosine similarity information 
of each support prototype and the query features is accumulated to generate a probability map.
Once the correspondence is established, the matched features will participate in the decoding phase.
With the recurrent prediction enhancement module (RPEM), the decoder can continuously enhance the semantic information of the probability map and gradually restore a more accurate segmentation result.

\subsection{Rich Prototype Generation}
\label{section:multiscale}

For generating a series of complementary prototypes, K-means clustering (KMC) \cite{PPNet} and superpixel-guided clustering (SGC) \cite{ASGNet}, two strategies with different scale-aware capabilities play a central role in the proposed rich prototype generation module (RPGM).
The internal structure of the RPGM is shown in Fig.~\ref{wang1}.

Specifically, we first apply KMC to compute a data partition $D=\{{D_{i}\}_{i=1}^{N_{k}}}$ and generate a set of prototypes ${{S}_{k}}=\{{{S}_{i}}\}_{i=1}^{{{N}_{k}}}$ by the average pooling as follows:
\begin{equation}\label{f1}
{{S}_{i}}=\frac{1}{|D_{i}|}\sum\limits_{j\in {{D}_{i}}}{F_{s}^{j}},
\end{equation}
where $|D_{i}|$ and ${F_{s}^{j}}$ represent the number of elements in $D_{i}$ and the support feature indexed by $D_{i}$, respectively.
In parallel, the collection of superpixel prototypes ${{S}_{s}}=\{{{S}_{r}}\}_{r=1}^{{{N}_{s}}}$ is produced from ${{F}_{s}}$ in the SGC branch:
\begin{equation}\label{f2}
{{S}_{r}}=\frac{\sum\nolimits_{p}{{{A}_{pr}}\cdot F_{s}^{p}}}{\sum\nolimits_{p}{{{A}_{pr}}}},
\end{equation}
where $A$ denotes the association mapping between each pixel $p$ and all superpixels.
The enhanced prototype ${{S}_{ks}}$ with diverse representations is the combination of the two sets of feature prototypes:
\begin{equation}\label{f3}
{{S}_{ks}}=C({{S}_{k}}, {{S}_{s}}),
\end{equation}
where $C(\cdot , \cdot)$ is the channel-wise concatenation operation. 

Then, we convey the integrated prototype ${{S}_{ks}}$ and the query feature ${{F}_{q}}$ to the GPA module for matching. 
In the GPA, the cosine similarity maps $\{{B}_{i}\}_{i=1}^{{N}_{k}+{N}_{s}}$ corresponding to each support prototype $S_{ks}^{i}$ and the query features are fed into a two-branch structure. 
In the first branch, the index value of the most relevant prototype at each pixel location $(x,y)$ is collected as the guide map $g$:
\begin{equation}\label{f4}
{{g}^{x,y}}=\underset{i\in \left\{ 0,...,{{N}_{k}}+{{N}_{s}} \right\}}{\mathop{\arg \max }}\,B_{i}^{x,y}.
\end{equation}
Based on $g$, the corresponding prototypes are gathered to form the guide feature map ${{F}_{g}}$.
In the second branch, all similarity maps are accumulated to obtain the probability map $P$:
\begin{equation}\label{f5}
P=\sum\limits_{i=1}^{{{N}_{k}}+{{N}_{s}}}{{{B}_{i}}}.
\end{equation}
More details about the SGC and the GPA can be found in~\cite{ASGNet}. 

Finally, the query feature ${{F}_{q}}$, the guide feature ${{F}_{g}}$ and the probability map $P$ are fed into the Q-FPN for further enhancement of feature information. The Q-FPN is a module that provides multi-scale input to the decoder~\cite{PFEnet}. According to~\cite{PFEnet}, the feature scales are chosen as ${\{60, 30,15,8\}}$.

\begin{figure*}[t!]
	\centering	
	\includegraphics[width=\linewidth]{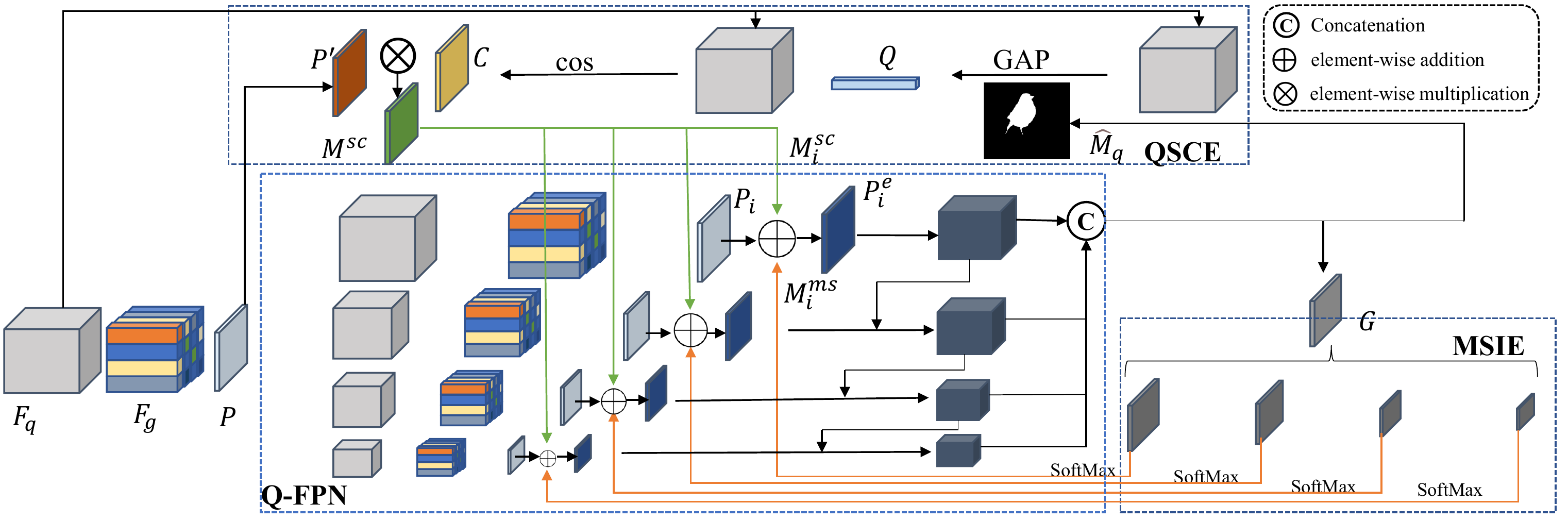}
	\caption{Visual illustration of our proposed RPEM.}
	\label{wang2}
\end{figure*}

\subsection{Recurrent Prediction Enhancement}
\label{sub:RME}

The recurrent mechanism has been applied to some segmentation methods~\cite{canet,rp}.
They rely heavily on the initial prediction to hard-compute correlations at the map-level by iterative operations and introduce more convolution parameters to optimize the prediction.
Different from them, we pass the initial prediction information into the GPA module, and update the probability map through a recurrent mechanism to refine the prediction in a soft way.
As shown in Fig.~\ref{wang2}, we propose a recurrent prediction enhancement module (RPEM), including two parts: the multi-scale iterative enhancement (MSIE) and the query self-contrast enhancement (QSCE). 

\noindent\textbf{Multi-scale Iterative Enhancement (MSIE).}
The multi-scale operation can provide multi-level spatial information for features. The output feature $G$ of the last layer is first processed with the adaptive average pooling operation to produce the feature set $\{{G}_{i}\}_{i=1}^{v}$ corresponding to $v$ different spatial sizes. 
And we perform the softmax operation for each ${{G}_{i}}$ to get a two-channel foreground-background probability map ${M}_{i}=\{{M}_{i}^c\}_{c=1}^{2}$.
To further enhance the information representation, a min-max normalization is introduced to process the foreground map $M_{i}^{1}$, as described below:
\begin{equation}\label{f6}
M_{i}^{1}=\frac{M_{i}^{1}-\min (M_{i}^{1})}{\max (M_{i}^{1})-\min (M_{i}^{1})+{{10}^{-7}}}.
\end{equation}
Finally, an affine transformation is applied to maintain interval consistency between $M_{i}^{1}$ and the probability map ${{P}_{i}}$:
\begin{equation}\label{f7}
M_{i}^{ms}=\alpha M_{i}^{1}-\beta,
\end{equation}
where ${\alpha}$ is set to 2 and ${\beta}$ is set to 1. 
$M_{i}^{ms}$ represents the additional information required by the probability map ${{P}_{i}}$. 

\noindent\textbf{Query Self-contrast Enhancement (QSCE).}
The pseudo mask ${\widehat{M}}_{q}$ is first generated from the two-channel map $G$ by the argmax operation. The foreground and background are represented by 1 and 0, respectively.
And the query prototype $Q$ can be obtained by averaging the masked query feature ${{F}_{q}}$:
\begin{equation}\label{f9}
Q=\frac{\sum\limits_{i=1}^{hw}{{{F}_{q}}(i)\odot \widehat{{{M}_{q}}}(i)}}{\sum\limits_{i=1}^{hw}{\widehat{{{M}_{q}}}(i)}},
\end{equation}
where ${\odot}$ is the broadcast element-wise multiplication.
The cosine distance between the query prototype and each query feature element is calculated as
\begin{equation}\label{f10}
{{C}}=\frac{Q\cdot F_{q}}{\left\| Q \right\|\cdot \left\| F_{q} \right\|},
\end{equation}
where ${C}$ is the self-contrast similarity information of the query feature. 
Following Eq. (\ref{f6}), the value of initial probability map $P$ is normalized to be between 0 and 1. By multiplying the normalized probability map ${P}'$ with $C$, we can generate the complementary information ${{M}^{sc}}$:
\begin{equation}\label{f12}
{{M}^{sc}}={P}'\cdot \frac{C+1}{2},
\end{equation}
where we convert the value on ${{C}}$ to between 0 and 1 to ensure that the same range of values and the operation make sense.
To maintain interval consistency, the same transformation is imposed on ${{M}^{sc}}$ as Eq. (\ref{f7}).
Similarly, we also perform a multi-scale operation on ${{M}^{sc}}$ to obtain $M_{i}^{sc}$.
%

Finally, we accumulate the additional information $M_{i}^{ms}$ and $M_{i}^{sc}$ obtained by the MSIE and the QSCE together to the probability map $P_{i}$ to generate a new enhanced probability map $P_{i}^{e}$ as
\begin{equation}\label{f13}
P_{i}^{e}={{P}_{i}}+M_{i}^{ms}+M_{i}^{sc}.
\end{equation}

\section{Experiments}
\label{section:experiments}
\subsection{Experimental Settings}
\label{sub:settings}

We evaluate the performance of our method on PASCAL-${{5}^{i}}$ and COCO-${{20}^{i}}$, two datasets that are frequently used to measure few-shot segmentation. PASCAL-${{5}^{i}}$ is composed of PASCAL VOC 2012~\cite{pascal} and additional annotations of the SBD dataset~\cite{sbd}, containing 20 categories. COCO-${{20}^{i}}$ is modified from the MSCOCO dataset~\cite{coco} and consists of 80 categories. For each dataset, all categories are divided equally into four folds and 
our method is evaluated with reference to the setting in ASGNet~\cite{ASGNet}.
Following~\cite{panet}, we use the widely adopted mean intersection over union (mIoU) as the evaluation metric.

\subsection{Implementation Details}
\label{sub:details}
All experiments are conducted on the PyTorch framework. Our approach is built on ASGNet~\cite{ASGNet}, and the backbone networks are ResNet-50~\cite{res} and ResNet-101~\cite{res}. The proposed model is trained at PASCAL-${{5}^{i}}$ for 200 epochs and at COCO-${{20}^{i}}$ for 50 epochs. 
Due to the limited computing resources, we choose a learning rate of 0.0025 and a batchsize of 4 for training on both datasets, but this does not prevent us from proving the effectiveness of our method. 
In this paper, we use the cross-entropy loss as supervision.
Based on the experience of PPNet~\cite{PPNet}, we set the number of clusters $N$ to 5 on each dataset, and please see ASGNet~\cite{ASGNet} for some other details. All of our experiments are conducted on an NVIDIA GTX 1080 GPU and an NVIDIA RTX 2080Ti GPU.
\begin{table}[!t]
  \centering
  \caption{
      Comparison with state-of-the-art methods on PASCAL-${{5}^{i}}$.
  }
  \begin{threeparttable}
    \resizebox{\linewidth}{!}{
      \begin{tabular}{|c|r||ccccc|ccccc|}
        \hline
        \rowcolor{mygray}
                                            &                                   & \multicolumn{5}{c|}{\textbf{1-shot}} & \multicolumn{5}{c|}{\textbf{5-shot}}                                                                                                                                         \\
        \rowcolor{mygray}
        \multirow{-2}{*}{\textbf{Backbone}} & \multirow{-2}{*}{\textbf{Method}} & \textbf{s-0}                         & \textbf{s-1}                         & \textbf{s-2}   & \textbf{s-3}   & \textbf{mean}  & \textbf{s-0}   & \textbf{s-1}   & \textbf{s-2}   & \textbf{s-3}   & \textbf{mean}  \\ \hline \hline
        \multirow{6}{*}{Res-50}             & CANet~\cite{canet}                & 52.50                                & 65.90                                & 51.30          & 51.90          & 55.40          & 55.50          & 67.80          & 51.90          & 53.20          & 57.10          \\
                                            & PFENet~\cite{PFEnet}              & 61.70                                & 69.50                                & 55.40          & 56.30          & 60.80          & 63.10          & 70.70          & 55.80          & 57.90          & 61.90          \\
                                            & SCL~\cite{SCL}                    & \textbf{63.00}                       & \textbf{70.00}                       & 56.50          & \textbf{57.70} & \textbf{61.80} & 64.50          & 70.90          & 57.30          & 58.70          & 62.90          \\
                                            & RePRI~\cite{RePRI}                & 59.80                                & 68.30                                & \textbf{62.10} & 48.50          & 59.70          & 64.60          & \textbf{71.40} & \textbf{71.10} & 59.30          & \textbf{66.60} \\
                                            & ASGNet~\cite{ASGNet}              & 58.84                                & 67.86                                & 56.79          & 53.66          & 59.29          & 63.66          & 70.55          & 64.17          & 57.38          & 63.94          \\
                                            & Ours                              & 60.95                                & 68.16                                & 59.87          & 55.12          & 61.03          & \textbf{65.16} & 70.71          & 69.51          & \textbf{60.30} & 66.42          \\ \hline
        \multirow{4}{*}{Res-101}            & PFENet~\cite{PFEnet}              & 60.50                                & 69.40                                & 54.40          & \textbf{55.90} & 60.10          & 62.80          & 70.40          & 54.90          & 57.60          & 61.40          \\
                                            & RePRI~\cite{RePRI}                & 59.60                                & 68.60                                & \textbf{62.20} & 47.20          & 59.40          & \textbf{66.20} & 71.40          & 67.00          & 57.70          & 65.60          \\
                                            & ASGNet~\cite{ASGNet}              & 59.84                                & 67.43                                & 55.59          & 54.39          & 59.31          & 64.55          & 71.32          & 64.24          & 57.33          & 64.36          \\
                                            & Ours                              & \textbf{60.80}                       & \textbf{69.83}                       & 58.19          & 54.93          & \textbf{60.94} & 64.86          & \textbf{72.72} & \textbf{71.11} & \textbf{59.99} & \textbf{67.17} \\ \hline
      \end{tabular}
    }
  \end{threeparttable}
  \label{tab1}
\end{table}

\begin{table}[!t]
  \centering
  \caption{Comparison with state-of-the-art methods on COCO-${{20}^{i}}$.}
  \begin{threeparttable}
  \resizebox{\linewidth}{!}{
      \begin{tabular}{|c|r||ccccc|ccccc|}
        \hline
        \rowcolor{mygray}
                                            &                                   & \multicolumn{5}{c|}{\textbf{1-shot}} & \multicolumn{5}{c|}{\textbf{5-shot}}                                                                                                                                         \\
        \rowcolor{mygray}
        \multirow{-2}{*}{\textbf{Backbone}} & \multirow{-2}{*}{\textbf{Method}} & \textbf{s-0}                         & \textbf{s-1}                         & \textbf{s-2}   & \textbf{s-3}   & \textbf{mean}  & \textbf{s-0}   & \textbf{s-1}   & \textbf{s-2}   & \textbf{s-3}   & \textbf{mean}  \\ \hline \hline
    \multirow{3}[2]{*}{Res-101} & DAN~\cite{DAN}   & -     & -     & -     & -     & 24.20  & -     & -     & -     & -     & 29.60  \bigstrut[t]\\
          & PFENet~\cite{PFEnet} & 34.30  & 33.00  & 32.30  & 30.10  & 32.40  & 38.50  & 38.60  & 38.20  & 34.30  & 37.40  \\
          & SCL~\cite{SCL}   & \textbf{36.40} & 38.60  & \textbf{37.50} & \textbf{35.40} & \textbf{37.00} & 38.90  & 40.50  & 41.50  & 38.70  & 39.00  \bigstrut[b]\\
    \hline
    \multirow{4}[2]{*}{Res-50} & RPMM~\cite{PMM}  & 29.53  & 36.82  & 28.94  & 27.02  & 30.58  & 33.82  & 41.96  & 32.99  & 33.33  & 35.52  \bigstrut[t]\\
          & RePRI~\cite{RePRI} & 31.20  & 38.10  & 33.30  & 33.00  & 34.00  & 38.50  & 46.20  & 40.00  & \textbf{43.60} & 42.10  \\
          & ASGNet~\cite{ASGNet} & 34.89  & 36.94  & 34.33  & 32.08  & 34.56  & 40.99  & \textbf{48.28} & 40.10  & 40.54  & 42.48  \\
          & Ours  & 36.21  & \textbf{39.47} & 33.97  & 34.07  & 35.93  & \textbf{43.05} & 47.91  & \textbf{42.71} & 41.24  & \textbf{43.73} \bigstrut[b]\\
    \hline
    \end{tabular}
  }
  \end{threeparttable}%
  \label{tab2}%
\end{table}%
\subsection{Results}
\label{sub:results}
\begin{figure}[!t]
  \centering
  \includegraphics[width=\linewidth]{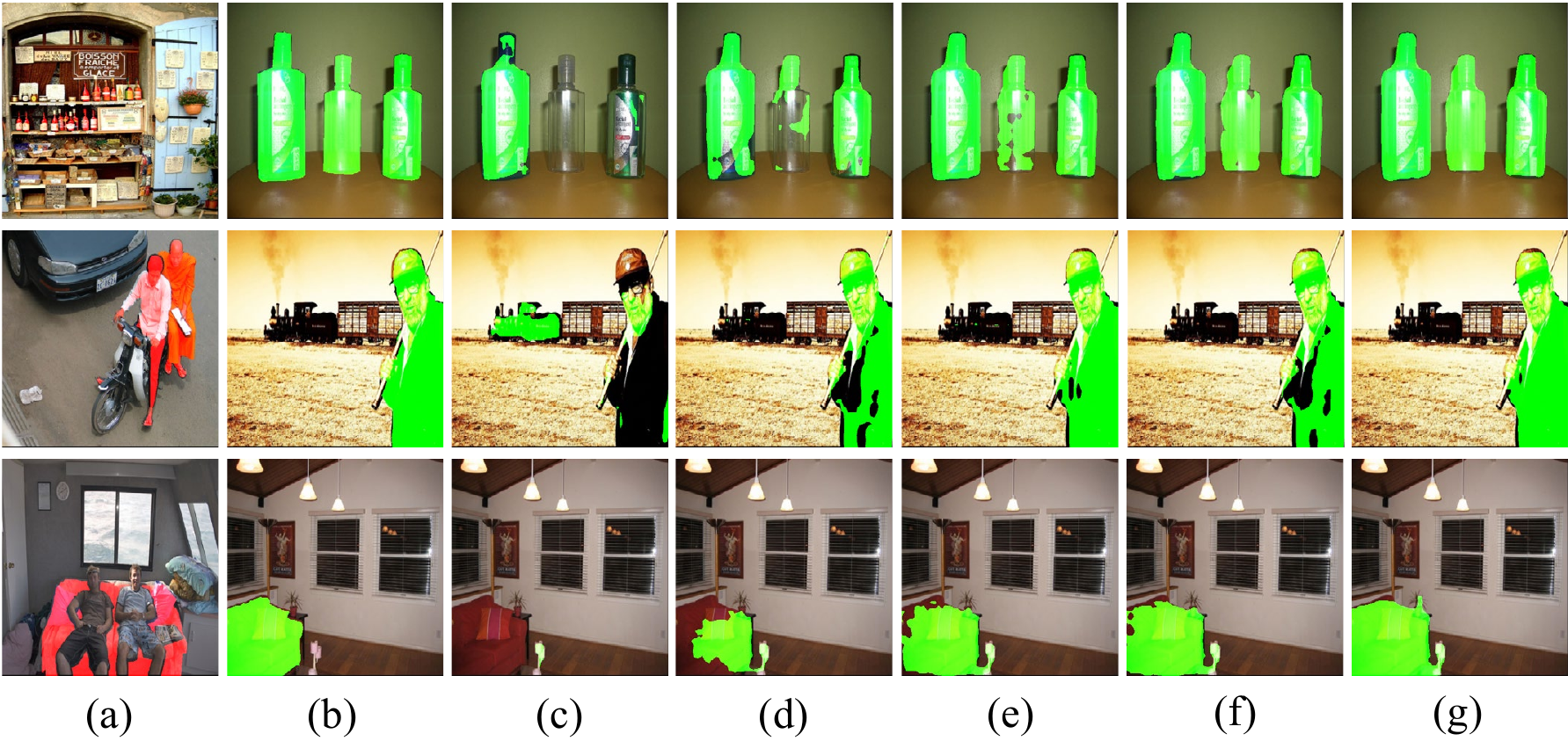}
  \caption{
    Qualitative comparison of different components on PASCAL-${{5}^{i}}$.
    (a) $I_{s}\&M_{s}$.
    (b) $I_{q}\&M_{q}$.
    (c) Baseline (ASGNet).
    (d) +RPGM.
    (e) +RPGM+MSIE.
    (f) +RPGM+QSCE.
    (g) +RPGM+RPEM.
  }
  \label{wang3}
\end{figure}

In Table \ref{tab1}, we compare the proposed method with other state-of-the-art methods on PASCAL-${{5}^{i}}$. Experiments 
show that our method significantly outperforms the baseline and achieves new state-of-the-art performance on the 5-shot setting. When using ResNet-101 as the backbone, our method improves 1.63$\%$ and 2.81$\%$ over ASGNet in 1-shot and 5-shot segmentation, respectively.
In Fig.~\ref{wang3}, we show some representative results for all our setups.
As can be seen from the first and third lines, our method can segment query images with different complexity very well even if the foreground of the support image is small.
In Table \ref{tab2}, 
our method outperforms 1.37$\%$ and 1.25$\%$ in terms of mIoU over the baseline ASGNet under the 1-shot and 5-shot setting on COCO-${{20}^{i}}$.

 \begin{table}[!t]
   \centering
   \caption{Ablation study of our RPGM and RPEM (MSIE and QSCE) on PASCAL-${{5}^{i}}$.}
   \small
   \begin{threeparttable}
     \resizebox{0.7\linewidth}{!}{
       \renewcommand\arraystretch{1}
       \begin{tabular}{|l||cc|cc|}
         \hline
         \rowcolor{mygray}
                                           & \multicolumn{2}{c|}{\textbf{Res-50}} & \multicolumn{2}{c|}{\textbf{Res-101}}                                     \\
         \rowcolor{mygray}
         \multirow{-2}{*}{\textbf{Method}} & \textbf{1-shot}                      & \textbf{5-shot}                       & \textbf{1-shot} & \textbf{5-shot} \\
         \hline \hline
         Baseline (ASGNet)                 & 59.29                                & 63.94                                 & 59.31           & 64.36           \\
         +RPGM                             & 60.13                                & 65.66                                 & 60.05           & 66.47           \\
         +RPGM+MSIE                        & 60.51                                & \textbf{66.43}                        & 60.73           & 67.13           \\
         +RPGM+QSCE                        & \textbf{61.07}                       & 65.72                                 & 60.64           & 66.73           \\
         +RPGM+RPEM (MSIE+QSCE)                        & 61.03                                & 66.42                                 & \textbf{60.94}  & \textbf{67.17}  \\
         \hline
       \end{tabular}
     }
   \end{threeparttable}
   \label{tab3}
 \end{table}

\subsection{Ablation Study}
\label{sub:Ablation}
In this subsection, we conduct extensive ablation studies 
on PASCAL-${{5}^{i}}$ to evaluate the effects of our proposed components.

\noindent\textbf{Effects of RPGM and RPEM.}
The comparison shown in Table \ref{tab3} validates the effectiveness of each proposed component. 
And the visualization in Fig.~\ref{wang3} also reflects that these modules can brings consistent improvements across different samples.
It can be seen that the RPGM can fully present the details of the supported images and the RPEM can refine the missing foreground in the query image segmentation.

\begin{figure}[!t]
  \centering
  \includegraphics[width=\linewidth]{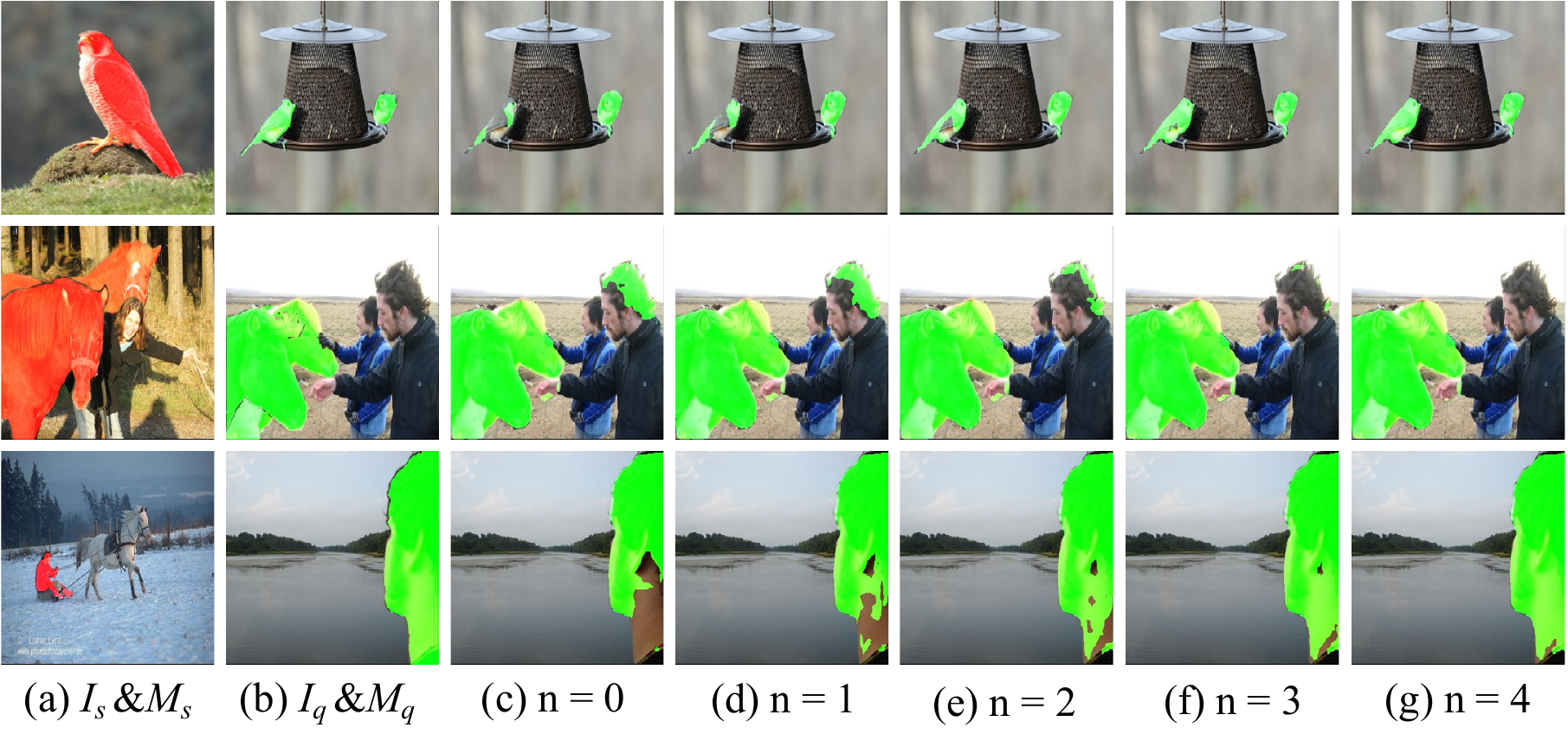}
  \caption{
    Predictions from different iterations.
  }
  \label{wang4}
\end{figure}

\begin{table}[!t]
\centering
\caption{Ablation study on the number of iterations $n$.}
\small
\begin{threeparttable}
 \resizebox{0.6\linewidth}{!}{
   \renewcommand\arraystretch{1}
   \begin{tabular}{|c||ccccc|}
     \hline
     \rowcolor{mygray}
     \textbf{$n$ iterations} & \textbf{s-0}   & \textbf{s-1}   & \textbf{s-2}   & \textbf{s-3}   & \textbf{mean}  \\
     \hline \hline
     0          & 60.06          & 68.16          & 58.43          & 53.88          & 60.13          \\
     2          & \textbf{61.03} & \textbf{68.30} & 59.47          & 54.89          & 60.92          \\
     4          & 60.95          & 68.16          & \textbf{59.87} & \textbf{55.12} & \textbf{61.03} \\
     6          & 60.34          & 67.58          & 59.86          & 54.87          & 60.66          \\
     \hline
   \end{tabular}
 }
\end{threeparttable}
\label{tab4}
\end{table}

\noindent\textbf{Number of Recurrent Prediction Iterations.}
To explore the optimal number of iterations, we conduct several controlled experiments and results are summarized in Table \ref{tab4}.
We observe that the model with the number of iterations ``$n=4$'' achieves the best mean performance.
In Fig.~\ref{wang4}, we also show the prediction of each iteration. 
It can be intuitively seen that as the iteration proceeds, the missing foreground parts of the query image are increasingly filled in while the misclassified background parts are gradually removed.

\section{Conclusion}
In this paper, we propose the novel rich prototype generation module (RPGM) and the recurrent prediction enhancement module (RPEM) for few-shot segmentation. 
The RPGM generates rich prototypes that compensate for the loss of detail 
when only a single type of prototypes is available.
The RPEM includes the multi-scale iterative enhancement (MSIE) and the query self-contrast enhancement (QSCE), which can be directly applied to 
gradually guide the probability map towards completeness and refine the segmentation map in the inference phase. 
Extensive experiments and ablation studies demonstrate the effectiveness of these proposed components, 
and our approach substantially improve the performance of the baseline on PASCAL-${{5}^{i}}$ and COCO-${{20}^{i}}$ datasets.

\noindent\textbf{Acknowledgements}
This work was supported by the National Natural Science Foundation of China \#62176039.
%
%
%
%

\bibliographystyle{splncs04}
\bibliography{ourbib}
\end{document}